\documentclass{article}

\usepackage{tccml_iclr2025_conference,times}

\usepackage[utf8]{inputenc} % allow utf-8 input
\usepackage[T1]{fontenc}    % use 8-bit T1 fonts
\usepackage{hyperref}       % hyperlinks
\usepackage{url}            % simple URL typesetting
\usepackage{booktabs}       % professional-quality tables
\usepackage{amsfonts}       % blackboard math symbols
\usepackage{nicefrac}       % compact symbols for 1/2, etc.
\usepackage{microtype} % microtypography
\usepackage{graphicx}
\usepackage{subcaption}
\usepackage{stfloats}
\usepackage{tabularx}

\DeclareUnicodeCharacter{2212}{-}

\title{Using street view imagery and deep generative modeling for estimating the health of urban forests}

\author{%
    Akshit Gupta\\
    TU Delft\\
    Delft, The Netherlands \\
    \texttt{a.gupta-5@tudelft.nl} \\
    \And
    Remko Uijlenhoet \\
    TU Delft\\
    Delft, The Netherlands \\
    \texttt{R.Uijlenhoet@tudelft.nl} \\
}
\iclrfinalcopy
\begin{document}

\maketitle

\begin{abstract}
Healthy urban forests comprising of diverse trees and shrubs play a crucial role in mitigating climate change. They provide several key advantages such as providing shade for energy conservation, and intercepting rainfall to reduce flood runoff and soil erosion. Traditional approaches for monitoring the health of urban forests require instrumented inspection techniques, often involving a high amount of human labor and subjective evaluations. As a result, they are not scalable for cities which lack extensive resources. Recent approaches involving multi-spectral imaging data based on terrestrial sensing and satellites, are constrained respectively with challenges related to dedicated deployments and limited spatial resolutions. In this work, we propose an alternative approach for monitoring the urban forests using simplified inputs: street view imagery, tree inventory data and meteorological conditions. We propose to use image-to-image translation networks to estimate two urban forest health parameters, namely, NDVI and CTD. Finally, we aim to compare the generated results with ground truth data using an onsite campaign utilizing handheld multi-spectral and thermal imaging sensors. With the advent and expansion of street view imagery platforms such as Google Street View and Mapillary, this approach should enable effective management of urban forests for the authorities in cities at scale.

\end{abstract}

\section{Introduction}

Urban forests comprising of trees and shrubs are a fundamental asset in urban cities whihc provide multiple benefits for mitigating climate change. These benefits range from intercepting rainfall to prevent water from directly flowing to impervious surfaces, thus reducing flood events, to reducing energy consumption by providing shade, thus reducing urban heat island effect and local temperatures (\cite{GREGORYMCPHERSON199241,Hobbie_2020}). However, with the changing urban climate, the existing urban forests need increased maintenance and more freuquent monitoring (\cite{climateChangeGreenery3}). This heightened attention is essential because the capacity of urban forests to deliver their diverse and beneficial ecosystem services is significantly compromised when these green spaces are unhealthy (\cite{urbanTreeManagementEcosystemDelivery}).

\begin{figure}[t]
    \centering
    \includegraphics[width=0.55\linewidth]{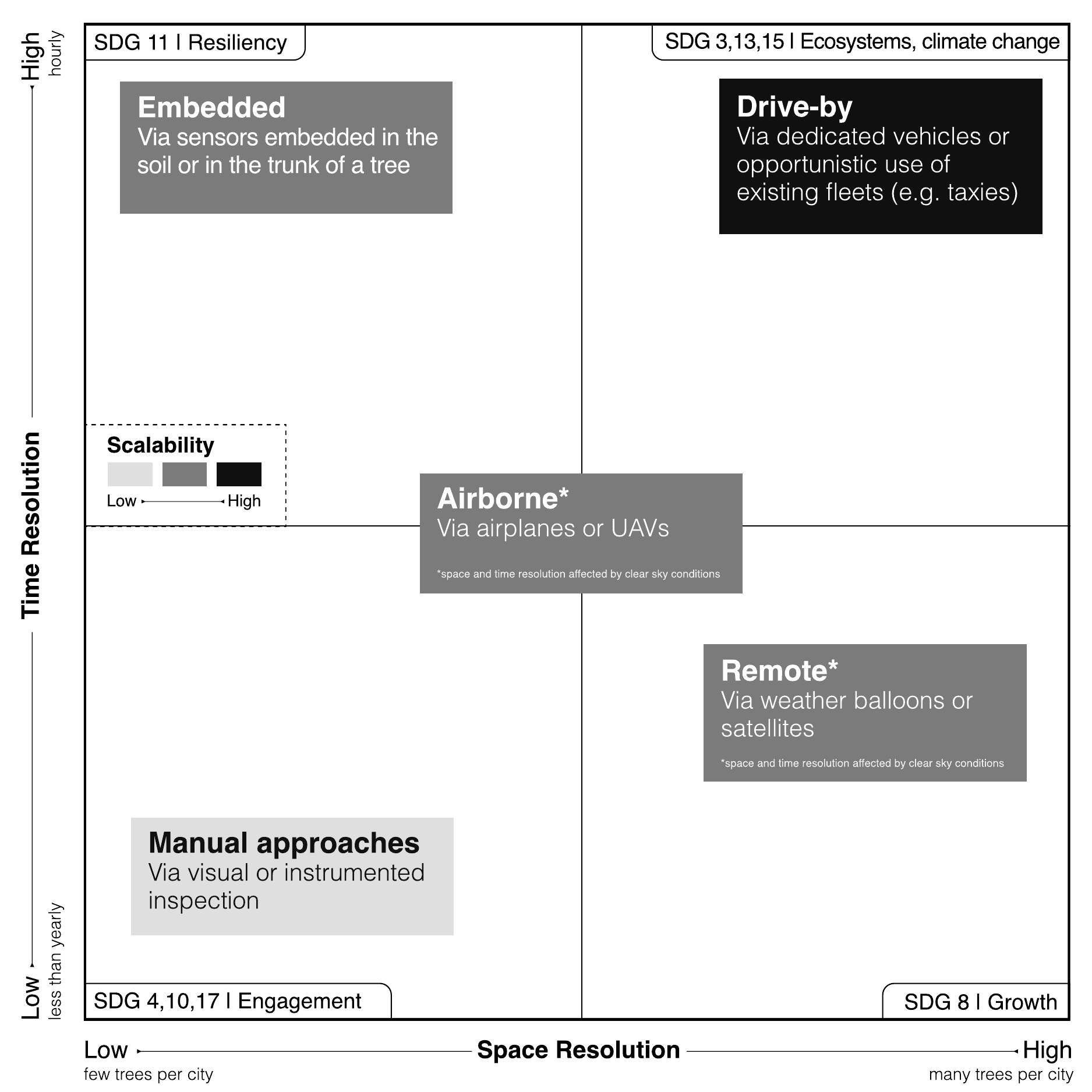}
    \caption{Current methods for monitoring health of urban forest and the scalability challenges associated with them due to spatial and temporal resolutions. \textit{Adapted from (\cite{akshitNatureSustain})} }    \label{fig:review}
\end{figure}

Currently, the health of urban forests is measured by arborists (or tree doctors) moving manually from one tree to next, which involves high human labor. This leads to infrequent health assessments with a temporal delay of two to four years. Recent approaches relying on technological developments (as shown in Figure \ref{fig:review}) such as LEO (low earth orbit) remote sensing data, embedded sensing can supplement traditional manual approaches (\cite{akshitNatureSustain}). However, these approaches are impeded respectively with challenges related to the spatial resolution of satellite imaging sensors and the scalability of embedded sensors. Recent vehicular mobile sensing approaches (shown as drive-by in Figure \ref{fig:review}) involve deployment of multiple imaging sensors on dedicated or opportunistic fleets such as cars, bikes etc, leading to high operational costs.

Both the technological approaches of remote sensing and drive-by sensing utilize multi-spectral or hyper-spectral imaging sensors deployed on a moving entity which can be satellites, air balloons, cars or bikes. These multi-spectral sensors generate multi-spectral images with additional channels (or spectral bands) compared to the standard RGB (Red, Green and Blue) channels associated with optical imagery. These additional channels (mainly comprising of NIR (near-infrared) and Thermal (or long-infrared)) are in-turn processed for calculating simple and reliable vegetation indexes which indicate urban tree health. For instance, GreenScan (\cite{akshitIEEESensors}) based on drive-by sensing utilizes two imaging sensors which concurrently capture Red, Green, Near Infrared images, and Thermal images to generate two tree health indexes namely NDVI (normalized difference vegetation index) and CTD (canopy temperature depression). 

As the measured health indexes have different threshold values based on the genus (and species) of the tree, the data related to genus (and species) ($I_{species}$) is indispensable. From the point of street level measurements, the value of light reflected, as is measured by the multi-spectral sensors depends on the input light radiation from the sun ($I_{radiation}$) and the solar orientation angle ($S_{angle}$) at the time of image capture. For measurement of CTD, air temperature ($T_{air}$) is essential. 

At the same time, image-to-image translation networks based on conditional GANs (generative adversarial networks) have been successfully utilized for generating output images with different characteristics (such as higher resolutions, additional channels) from input images in various application domains. Namely, in the remote sensing domain, \cite{condGANNIRfromRGB} utilize a conditional GAN to generate NIR channel data as output using RGB images as input for the Sentinel-2 satellite. In the agriculture domain, \cite{applicationPix2Pix} utilize a conditional GAN based on Pix2Pix architecture (\cite{pix2pix}) to generate near-infrared and red-edge channel data from RGB optical images captured using drones. Finally, in the urban planning domain, \cite{instantInfrared} utilize RGB optical images captured from street view to generate thermal imaging data and estimate urban facade temperatures.

\section{Proposed Methodology}

\begin{figure}[th]
    \centering
    \includegraphics[width=0.65\linewidth]{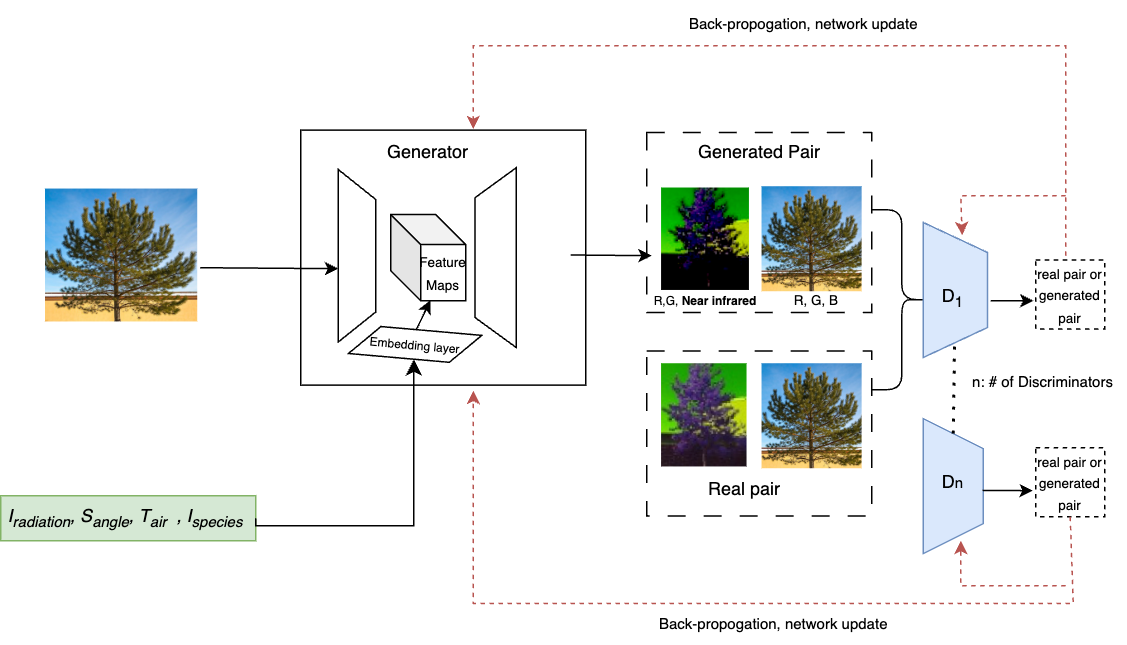}
    \caption{Adding the meteorological variables as an embedding layer in Pix2Pix conditional GAN architecture to generate near-infrared from standard RGB images. The same methodology is used to generate thermal images.}    \label{fig:architecuture}
\end{figure}

\begin{figure}[th]
    \centering
    \includegraphics[width=0.65\linewidth]{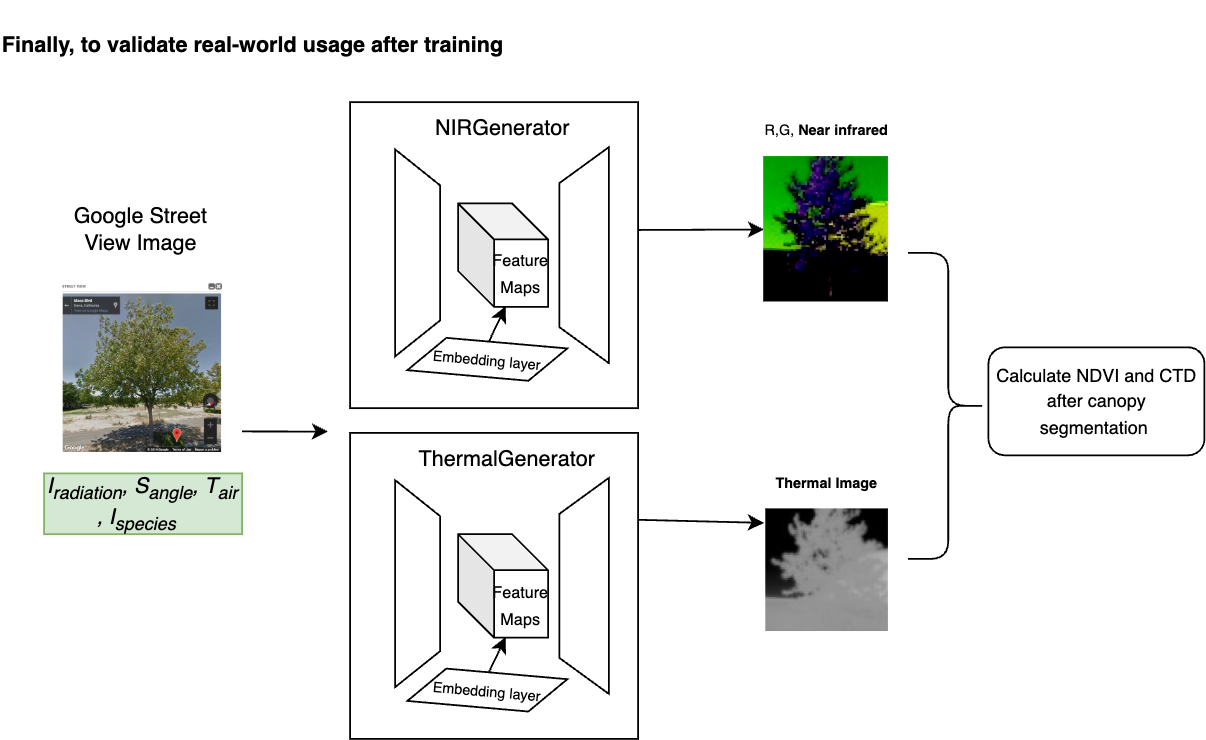}
    \caption{Using street view images from Google Street view to generate R,G, near infrared and thermal images, though previously trained models. The CTD and NDVI is calculated after segmentation of canopy as performed in (\cite{akshitIEEESensors} }    \label{fig:testTime}
\end{figure}

In this work, we propose to train a predictive model which generates both near-infrared and thermal images from RGB images using Pix2Pix (\cite{pix2pix}) architecture based on conditional GANs.
The Pix2Pix architecture is modified with input variables namely, ($I_{radiation}$), ($S_{angle}$), ($I_{species}$) fed as an embedding vector in the Generator stage of Pix2Pix. This is shown in Figure \ref{fig:architecuture}. Finally, the generated near-infrared and thermal images are in-turn used after canopy segmentation to estimate two tree health indexes namely NDVI and CTD (as shown in Figure \ref{fig:testTime}). By training a model on input paired imaging dataset, we bypass the need for capturing near-infrared and thermal imaging data, instead relying on non-linear correlations and embedding features learnt by the proposed model. This decreases the operational costs associated with the multiple imaging sensors utilized in the technical approaches of vehicular mobile and remote sensing.

For real world testing, our pre-trained model is applied to street view images obtained from the Mapillary platform as input data, to generate near-infrared and thermal images which are subsequently utilized to estimate NDVI and CTD. Finally, the generated NDVI and CTD values are compared with an on-site campaign with professional equipments comprising of a multispectral camera (MAPIR Survey 3W) and a thermal camera (FLIR A50).

% \todo{See Figure 2 for a detailed overview of the our proposed methodology}
\subsection{Datasets}
We propose to utilize the dataset from the study (\cite{akshitIEEESensors}) which includes pairs of RGB, near infrared and thermal images along with the location metadata. This data is supplemented with additional datasets from the studies of \cite{NIRVisibleDataset} and \cite{IRVisibleDataset}, which are first processed to segment our objects of interest (urban forest comprising of trees, shrubs etc.). The incident sun radiation ($I_{radiation}$), solar orientation angle ($S_{angle}$) and the air temperature ($T_{air}$) at the location and time of input image capture can be obtained though city-wide weather station services and do not need to be highly localized. The data related to tree genus ($I_{species}$) will be obtained from tree inventory datasets.

\section{Pathway to impact}
This work aims at advancing the development of urban forests monitoring which facilitate multiple ecosystem services which contributes towards energy conservation, reduction of extreme events, in addition to minor carbon sequestration, thus, mitigating the effects of urban climate change. Further, we hope to provide easy and responsible access benefiting a wider community including urban forestry companies and broader academic research areas, including urban planning and climate sciences. To ensure a long-term impact, we will keep this work well documented with easy guidelines for collaboration and responsible model access.

\bibliographystyle{unsrtnat}
\bibliography{references}
\end{document}